\PassOptionsToPackage{dvipsnames}{xcolor}
\documentclass[11pt]{article}
\usepackage[final]{emnlp2021}

\usepackage{times}
\usepackage{latexsym}

\usepackage{tikz}
\usetikzlibrary{positioning,arrows}

\usepackage{balance}

\usepackage{booktabs}
\usepackage{adjustbox}
\usepackage{graphicx}
\usepackage{pifont}
\newcommand{\cmark}{\ding{51}}%
\usepackage{tabulary}

\usepackage{microtype}

\usepackage{amsmath,amsfonts,bm, bbm}

\def\eqref#1{equation~\ref{#1}}

\def\1{\bm{1}}

\def\rvc{{\mathbf{c}}}

\def\rvw{{\mathbf{w}}}

\def\va{{\bm{a}}}

\def\vm{{\bm{m}}}

\def\mM{{\bm{M}}}

\DeclareMathAlphabet{\mathsfit}{\encodingdefault}{\sfdefault}{m}{sl}
\SetMathAlphabet{\mathsfit}{bold}{\encodingdefault}{\sfdefault}{bx}{n}

\newcommand{\sigmoid}{\sigma}

\usepackage{mdframed}

\usepackage{accents}
\newlength{\dhatheight}

\usepackage{lipsum}

\title{Math Word Problem Generation with \\ Mathematical Consistency and Problem Context Constraints}

\author{Zichao Wang $^{1}$ 
\\\And
  Andrew S. Lan $^{2}$ \\[5pt]
  $^1$ Rice University \hspace{15pt}
  $^2$ University of Massachusetts Amherst \hspace{15pt}$^3$ OpenStax\\
  \texttt{\{zw16,richb\}@rice.edu},\hspace{5pt} \texttt{andrewlan@cs.umass.edu} \\\And 
  Richard G.~Baraniuk $^{1,3}$
  }

\date{}

\begin{document}
\maketitle
\begin{abstract}
We study the problem of generating arithmetic math word problems (MWPs) given a math equation that specifies the mathematical computation and a context that specifies the problem scenario. 
Existing approaches are prone to generating MWPs that are either mathematically invalid or have unsatisfactory language quality. 
They also either ignore the context or require manual specification of a problem template, which compromises the diversity of the generated MWPs.
In this paper, we develop a novel MWP generation approach that leverages i) pre-trained language models and a context keyword selection model to improve the language quality of the generated MWPs and ii) an equation consistency constraint for math equations to improve the mathematical validity of the generated MWPs. 
Extensive quantitative and qualitative experiments on three real-world MWP datasets demonstrate the superior performance of our approach compared to various baselines. 

\end{abstract}

\section{Introduction}

Math word problems (MWPs) are an important type of educational resource that help assess and improve students' proficiency in various mathematical concepts and skills~\cite{Walkington2013,Verschaffel2020}. An MWP usually has a corresponding underlying math equation that students will need to identify by parsing the problem and then solve the problem using this equation.
An MWP is usually also associated with a ``context'', i.e., the (often real-world) scenario that the math equation is grounded in, expressed in the question's text.
The equation associated with an MWP is often exact and explicit, while the context of the MWP is more subtle and implicit. 
It is not immediately clear how the context information can be extracted or represented. 
Table~\ref{tab:data-example} shows an example of an MWP and its associated equation.

\begin{table}[]
\caption{An examples of MWP and its underlying equation. See Table~\ref{tab:data} for more information on the datasets.}
\vspace{-5pt}
\label{tab:data-example}
\centering
\resizebox{\columnwidth}{!}{%
\begin{tabular}{p{1.34\linewidth}}
\toprule
{\bf MWP}: Joan found 70 seashells on the beach . She gave Sam some of her seashells . She has 27 seashells . How many seashells did she give to Sam ?                                                                               \\\hline
{\bf Equation}: \texttt{x = (70 - 27) }                                                           
\\\bottomrule
\end{tabular}
}
\vspace{-10pt}
\end{table}

In this work, we study the problem of \emph{automatically generating MWPs} from equations and context, which is important for three reasons. 
First, an automatic MWP generation method can aid instructors and content designers in authoring MWP questions, accelerating the (often costly and labor-intensive) MWP production process.
Second, an automated MWP generation method can generate MWPs tailored to each student's background and interests, providing students with a personalized learning experience~\cite{Walkington2013} that often leads to better engagement and improved learning outcomes~\cite{Patricia2000,doi:10.1177/0963721412443552,Karpicke966,Koedinger:2015:LSS:2724660.2724681,Kovacs:2016:EIQ:2876034.2876041,Rohrer2010}. 
Third, an automated MWP generation method can potentially help instructors promote academic honesty among students. 
While new technologies create new learning opportunities, instructors have growing concerns of technologies that enable students to easily search for answers online without actually solving problems on their own~\cite{84f6c08b47844b54868caf82c625fc66,Lancaster2021}. 
Automatically generated MWPs that are unique and previously unseen yet preserve the underlying math components can potentially reduce plagiarism.

In addition to its educational utility, MWP generation is also technically challenging and interesting. An important consideration for MWP generation is {\em controllability}: in practice, human instructors or content designers often have clear preferences in the type of MWPs they want to use. 
Therefore, an MWP generation method should be able to generate MWPs that are of high language quality and are textually and mathematically consistent with the given equations and contexts. %
To date, there exist limited literature on MWP generation. Most prior works focus on automatically answering MWPs, e.g.,~\cite{li-etal-2019-modeling,li-etal-2020-graph-tree,qin-etal-2020-semantically,shi-etal-2015-automatically,wang-etal-2018-translating,roy-roth-2015-solving,wu-etal-2020-knowledge} instead of generating them~\cite{Nandhini2011,DBLP:conf/aaaifs/Williams11,10.5555/2832249.2832302,deane2003automatic}. 
Existing MWP generation methods also often generate MWPs that either are of unsatisfactory language quality or fail to preserve information on math equations and contexts that need to be embedded in them. See Section~\ref{sec:rw} for a detailed discussion.

\subsection{Contributions}
In this work, we take a step towards controllable generation of mathematically consistent MWPs with high language quality. 
Our approach leverages a pre-trained language model (LM) as the base model for improved language quality. The input to the LM is an equation and a context, from which the LM generates an MWP. On top of that, we introduce 2 components that impose constraints on the mathematical and contextual content of the generated MWP.
First, to improve mathematical consistency and control over equations, we introduce an equation consistency constraint, which encourages the generated MWP to contain the exact same equation as the one used to generate it. 
Second, to improve control over contexts, we introduce a context selection model that automatically extracts context from an MWP. 
Quantitative and qualitative experiments on real-world MWP datasets show that our approach (often significantly) outperforms various baselines on various language quality and math equation accuracy metrics.

\begin{figure}
    \centering
    \includegraphics[width=0.95\linewidth]{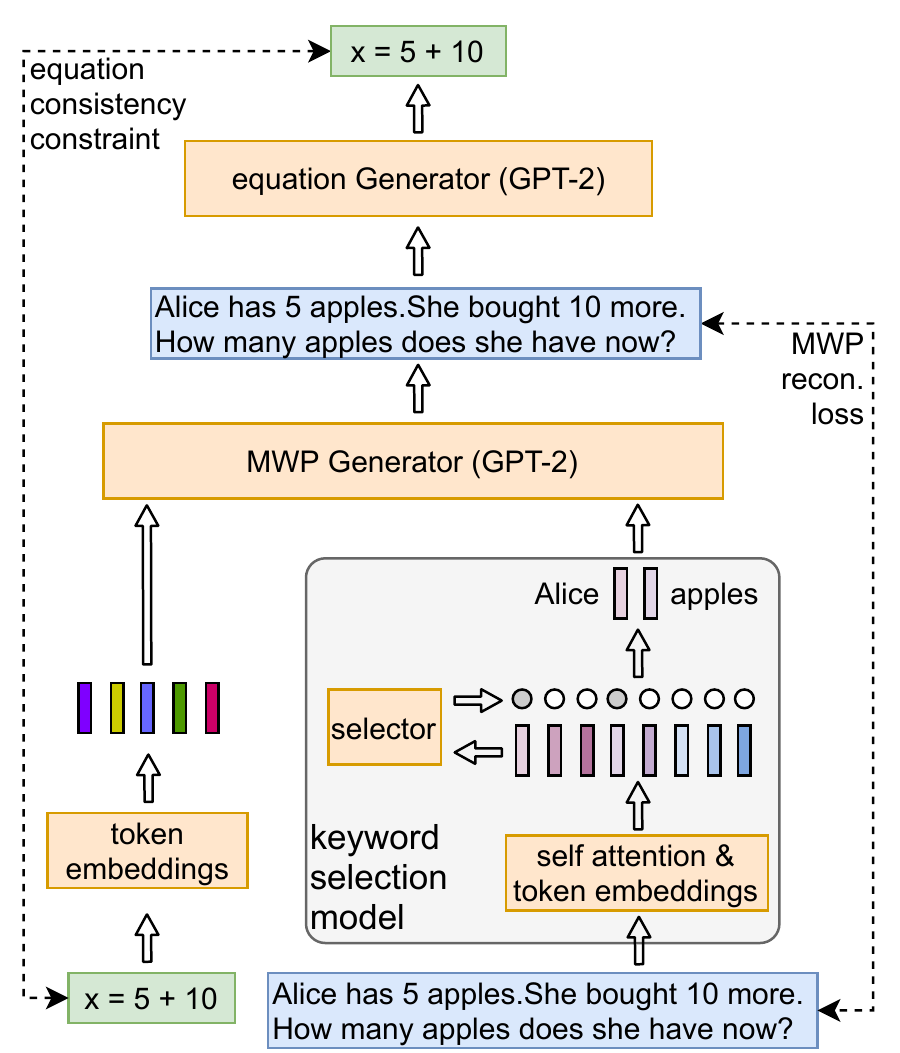}
\caption{An illustration of our MWP generation approach and its key components.}
\label{fig:method}
\vspace{-5pt}
\end{figure}

\section{Methodology}
We formulate the task of controllable MWP generation as a conditional generation problem. In this paper, we work with datasets $\mathcal{D} = \{ (M_i, E_i) \}_{i=1}^N$ in the form of $N$ (MWP, equation) pairs  where $M_i$ and $E_i$ represent MWP and its associated equation, respectively. In the remainder of the paper, we will remove the data point index to simplify notation. This setup assumes each MWP in our dataset is labeled with an underlying equation but its context is unknown. Then, the MWP generation process can be described as
\begin{align}
     M & \sim \mathbb{E}_{E\sim \mathcal{D}}[p_\Theta( M | E)] \nonumber \\
     & =  \mathbb{E}_{E\sim \mathcal{D},\, \rvc\sim p(\rvc|E,M)}[p_\Theta( M | E, \rvc )] \,,\label{eq:gen-model-original} 
\end{align}
where $M = \{m_1,\ldots,m_T\}$ represents the MWP as a sequence of $T$ tokens (e.g., words or wordpieces~\cite{NIPS2017_3f5ee243,radford2019language}). 
$E$ and $\rvc$ are the controllable elements, where $\rvc$ represents a problem context.
$p_\Theta$ is the MWP generative model parametrized by a set of parameters $\Theta$.

In this work, we use a pre-trained language model (LM) as the generative model $p_\Theta$, similar to the setup in~\cite{keskarCTRL2019}.
We choose LMs over other approaches such as sequence-to-sequence (seq2seq) models because they can be pre-trained on web-scale text corpora. Pre-trained LMs thus often generate high-quality text and generalizes well to out-of-domain words not present in the training data.
Under an LM, we can further decompose Eq.~\ref{eq:gen-model-original} into
\begin{align}
    p_\Theta( M | E, \rvc ) = \prod_{t=1}^T p_\Theta (m | E, \rvc, \{m_s\}_{s=1}^{t-1})\,.\label{eq:decompose}
\end{align}
To train $p_\Theta$ via fine-tuning the LM, we use the usual negative log-likelihood objective:
\begin{align}
    \mathcal{L}_{LM} = \sum_{t=1}^T -{\rm log}\, p_\Theta(m_t | E, \rvc, \{m_s\}_{s=1}^{t-1} )\,.\label{eq:loss-lm}
\end{align}

The above training objective serves as a proxy that optimizes for language quality. However,
it alone is unsatisfactory in 2 ways.
First, there is no guarantee that the generated MWP is mathematically valid; even if it is, its solution may correspond to an equation that is different from the input equation~\cite{zhou-huang-2019-towards}. 
Second, while the context $\rvc$ can be manually specified, i.e., as a set of keywords, it is unobserved during training and needs to be inferred from data through the costly-to-compute posterior distribution. 
In the remainder of this section, we introduce our novel approach to tackle these challenges. 
We first describe our equation consistency constraint 
that improves the generated MWP's mathematical consistency 
and then detail our context selection method that learns to extract the context in the form of a set of keywords from an MWP. 
Figure~\ref{fig:method} provides a high-level overview of our overall approach.

\subsection{Equation Consistency}
\label{sec:eq}

We propose an equation consistency constraint to promote the generated MWP to correspond to an equation that is the same as the input equation used to generate the MWP. 

To formulate this constraint, we need a model to parse an equation given an MWP, i.e., a mwp2eq model, and a loss function, which we call $\mathcal{L}_{eq}$. The mwp2eq generative process can be written as
\begin{align}
    & E'\sim  \mathbb{E}_{M'\sim p_\Phi(M|E)} [ p_\Phi ( E | M' ) ]\,,\nonumber
\end{align}
where $p_\Phi$ is the mwp2eq model, $E$ represents an equation, and $M'$ represents the generated MWP. 
Here, we treat the equation as a sequence of math symbols $e_t$, making it appropriate for sequential processing.
Specifically, we treat each variable (e.g., \texttt{x}, \texttt{y}), math operator (e.g., \texttt{=}, \texttt{$\times$}, \texttt{+}), and numeric value (e.g., integers, fractions, and decimal numbers) as a single math symbol.
Therefore, we can decompose $p_\Phi (E, M')$ similar to Eq.~\ref{eq:decompose}.
There are ways to represent math equations other than a sequence of symbols, such as symbolic trees~\cite{operator-tree-2,tangent-s,TangentCFT}; finding ways to make them compatible to LMs is left for future work. 
Similar to $\mathcal{L}_{\rm LM}$, we minimize a negative log-likelihood loss that uses the input equation $E$ as the ground truth for the equation $E'$ parsed from $M'$:
\begin{align}
    \mathcal{L}_{\rm eq} &= \sum_{t=1}^T -{\rm log}\,p_\Phi ( e_t | M', e_1,\ldots,e_{t-1} )\,. \label{eq:eq}
\end{align}
This constraint is reminiscent of the idea of ``cycle consistency'' that have found success in image and text style transfer~\cite{8237506,NIPS2017_2d2c8394}, question answering~\cite{yang-etal-2017-semi,1706.01450}, and disentangled representation learning~\cite{Jha_2018_ECCV}. 

\paragraph{Gumbel-Softmax Relaxation.}
To back-propagate loss to $p_\Theta$ and compute gradient for $\Theta$, we need the loss $\mathcal{L}_{\rm eq}$ be differentiable with respect to $\Theta$. The challenge here is that $M'$ is sampled from $p_\Theta$ and that this discrete sampling process is non-differentiable, preventing gradient propagation~\cite{nie2018relgan}. To tackle this challenge, we resort to the Gumbel-softmax relaxation~\cite{gumbel1,gumbel2} of the discrete sampling process $m_t\sim p_\Theta$. Details are deferred to the Supplementary Materials.

We remark that the gradient derived under the Gumbel-softmax relaxation is a biased but low-variance estimate of the true gradient~\cite{gumbel1,gumbel2}. The low-variance property makes it more attractive for real applications than other unbiased but high-variance estimators such as REINFORCE~\cite{10.1007/BF00992696}.
We refer to~\cite{gumbel1,gumbel2} for more details on the Gumbel-softmax method. In addition, while one can also use deterministic relaxation such as softmax, Gumbel-softmax injects stochastic noise during the training process, which regularizes the model and potentially improves performance; See an empirical comparison in Table~\ref{tab:ablation}.

\subsection{Context Selection}
In practice, we do not have access to the contexts $\rvc$ during training since they are not specified for real-world MWPs. 
Therefore, we need ways to specify the context for the MWP generative process. 
Existing methods characterize context as a ``bag-of-keywords'', using heuristic methods such as TF-IDF weights to select a subset of tokens as ``keywords'' from an MWP as its context. 
These methods are simple but lack flexibility: they either require one to specify the number of tokens to use for each MWP or heuristically select only certain types of tokens (e.g., nouns and pronouns)~\cite{zhou-huang-2019-towards,2010.06196}. 

In this work, we adopt this ``bag-of-tokens'' characterization of context, which fits well into LMs, but instead \emph{learn} a context (token) selection method from data.
To do so, we interpret $\rvc$ as a ``context keyword selection'' variable, i.e., a binary random vector whose dimension is the number of tokens in the vocabulary. 
Each entry $c^{(i)}$ in $\rvc$ is an i.i.d.\ Bernoulli random variable with prior probability $\rho$, i.e., $p_c(c^{(i)}=1) = \rho$. Thus, $\rvc$ acts as a selector that chooses appropriate context tokens from the entire vocabulary. 
To circumvent the intractable posterior $p(\rvc | E, M)$, we resort to the auto-encoding variational Bayes (VAE) paradigm~\cite{1312.6114}, similar to~\cite{shen-etal-2019-select}. 
Under the VAE setup, we select a set of tokens conditioned on the MWP as $\rvc \sim q_\Psi (M)$ where $q_\Psi(M)$ is a proposal distribution, i.e., the keyword selection model.

\paragraph{Context Keyword Selection Model.}
Given an MWP, we first compute the contextualized embeddings of each token 
using a simple linear self-attention method as 
\begin{align}
\widetilde{\vm}_t &= \mM \va_t\,,\quad \va_t = {\rm softmax}\left( \frac{\mM^\top\vm_t}{\sqrt{D}} \right), \nonumber
\end{align}
where $\mM = [\vm_1,\ldots,\vm_T] \in\mathbb{R}^{D\times T}$ is the matrix with all token embeddings and $D$ is the embedding dimension. 
The $\sqrt{D}$ term is added for numerical stability~\cite{NIPS2017_3f5ee243}. 
Then, we compute $q_{\Psi}^{(i)}(M)$, the probability that each word in the vocabulary is selected as a context keyword, with a single projection layer with Sigmoid activation
\begin{align}
    q_{\Psi}^{(i)} (M) = \sigmoid( \rvw^\top \widetilde{\vm}_t + b )\, \mathbf{1}_{\{V^{(i)}\in M\}}\,,\label{eq:c}
\end{align}
where $\rvw$ and $b$ are part of the model parameters $\Psi$. 
The indicator function at the end ensures that only tokens that appear in $M$ can be selected as context keywords. 
In practice, we also mask out stopwords and punctuation; these steps ensure that the context selector selects keywords that are relevant to MWPs and are not too generic.

\paragraph{Optimization Objective.}
Under the VAE paradigm, we optimize the keyword selection model using the so-called evidence lower bound (ELBO):
\begin{align}
    &\mathcal{L}_{\rm VAE} = \mathcal{L}_{\rm LM} + \beta \mathcal{L}_{\rm c}\,,\,\, \label{eq:context}
\end{align}
where $\mathcal{L}_{\rm c} = {\rm KL} (q_\Psi \| p_c)\, \mathbf{1}_{\{V^{(i)}\in M\}}$ and the Kullback-Leibler divergence term can be computed analytically thanks to our Bernoulli parametrization.
$\mathcal{L}_{\rm c}$ can be interpreted as a context constraint that prevents the keyword selection model from choosing too many keywords. The hyperparameter $\beta$ and prior $\rho$ controls the strength of this constraint. Because $\rvc$ is discrete and its sampling process is also non-differentiable, we use the straight-through estimator of the gradient~\cite{1308.3432} for $\Theta$ involved in $\mathcal{L}_{\rm LM}$ in Eq.~\ref{eq:context}.

\subsection{Training}
We train (fine-tune) the LM, the mwp2eq model, and the keyword selection model jointly. The mwp2eq model and keyword selection model are optimized using their respective objectives defined in Eqs.~\ref{eq:eq} and~\ref{eq:context}. The overall objective for the MWP generative model $p_\Theta$ is
\begin{align}
\mathcal{L} = \mathcal{L}_{\rm LM} + \alpha \mathcal{L}_{\rm eq} + \beta \mathcal{L}_{\rm c}\,,\nonumber
\end{align}
where $\alpha>0$ and $\beta>0$ are hyperparameters that balance these constraint terms.

\section{Experiments}
We now perform a series of experiments to validate the effectiveness of our proposed MWP generation approach. Quantitatively, we compare our approach to several baselines on various automated language quality and mathematical consistency metrics. Qualitatively, we showcase the capability of our approach in generating controllable, high-quality MWPs.

\begin{table}[t]
\caption{Summary statistics of datasets.}
\vspace{-5pt}
\label{tab:data}
\centering
\begin{adjustbox}{max width=\linewidth}
\begin{tabular}{@{}lccccc@{}}
\toprule
{\bf Dataset}    & {\bf \#MWPs} & {\bf avg \#words per MWP} & {\bf avg \#symbols per eq} \\ \midrule
{\bf arithmetic} & 1,492   & 29.89   & 8.05       \\
{\bf MAWPS}      & 2,373   & 31.25   & 8.16       \\
{\bf Math23K}    & 23,162  & 35.23   & 8.78       \\ \bottomrule
\end{tabular}
\end{adjustbox}
\vspace{-5pt}
\end{table}

\begin{table*}[pt!]
\caption{A comparison of language quality and mathematical validity for MWPs generated by our method to various baselines. Numbers in brackets indicate the accuracy of the mwp2eq model trained on each dataset, which is an upper bound on the performance under the ACC-eq metric.}
\vspace{-5pt}
\label{tab:quant-results}
\begin{adjustbox}{max width=\textwidth}
\begin{tabular}{@{}lcccccccccccc@{}}
\toprule
                      & \multicolumn{4}{c}{\bf Arithmetic}                                                                                                & \multicolumn{4}{c}{\bf MAWPS}                                                                                                     & \multicolumn{4}{c}{\bf Math23K}                                                                                                   \\ \midrule
                      & BLEU-4                        & METEOR                        & ROUGE-L                       & ACC-eq                        & BLEU-4                        & METEOR                        & ROUGE-L                       & ACC-eq                        & BLEU-4                        & METEOR                        & ROUGE-L                       & ACC-eq                        \\
& & & & (0.769) & & & & (0.755) & & & & (0.672) \\
                      \cmidrule(lr){2-5} \cmidrule(lr){6-9} \cmidrule(lr){10-13}
seq2seq-rnn  & 0.075                              & 0.152                              & 0.311                               & 0.413                              & 0.153                              & 0.175                               & 0.362                              & 0.472                              & 0.196                               &  0.234                             & 0.444                              & 0.390                              \\
\hspace{15pt} + GloVe  & {\bf 0.351}            & 0.310          & 0.555    & 0.399                          & 0.592          & 0.412         &  0.705                              & {0.585}                              & 0.275          & 0.277         & 0.507                               & 0.438                                                             \\
seq2seq-tf & 0.339          & 0.298         &  0.524      & 0.405                        & 0.554          & 0.387         & 0.663                               & {\bf 0.588}                               & 0.301          & 0.294         & 0.524                               &  {\bf 0.509}                                                          \\\midrule
GPT                   & 0.237 & 0.248 & 0.455 & 0.401 & 0.368 & 0.294 & 0.538 & 0.532 &  0.282          & 0.297         & 0.512  & 0.477 \\
GPT-pre        & 0.316                         & {\bf 0.322}                         & 0.554                         & 0.403                         & 0.504                         & 0.391                         & 0.664                         & 0.512                         & 0.325                         & {\bf 0.333}                         & {\bf 0.548}                         & 0.498                         \\
ours                  & {0.338}                & {\bf 0.322}                         & \textbf{0.567}                & {\bf 0.453}                & \textbf{0.596}                & \textbf{0.427}                & \textbf{0.715}                & {0.557}                & \textbf{0.329}                & 0.328                 & {0.544}                         & {0.505}                \\ \bottomrule
\end{tabular}
\end{adjustbox}
\end{table*}

\begin{table}[]
\caption{\% of generated MWPs that are {\em not} present in the training data. Our approach generates novel MWPs not seen in the training data most of the time while seq2seq-tf may simply memorize the training data.}
\label{tab:unique}
\vspace{-5pt}
\centering
\begin{adjustbox}{max width=0.8\linewidth}
\begin{tabular}{@{}lccc@{}}
\toprule
                  & \textbf{Arithmetic} & \textbf{MAWPS} & \textbf{Math23K} \\ \midrule
seq2seq-tf        & 6.24\%              & 2.49\%         & 38.88\%          \\
\textbf{ours}     & {\bf 94.90}\%             & {\bf 63.77}\%        & \textbf{95.72\%} \\ \bottomrule
\end{tabular}
\end{adjustbox}
\vspace{-5pt}
\end{table}

\paragraph{Datasets.}
\label{sec:data}
We focus on MWP datasets in which each MWP is associated with a single equation and each equation contains a single unknown variable. Therefore, we consider three such MWP datasets including {\bf Arithmetic}~\cite{hosseini-etal-2014-learning}, {\bf MAWPS}~\cite{koncel-kedziorski-etal-2016-mawps}, and {\bf Math23K}~\cite{wang-etal-2017-deep-neural}. 
Table~\ref{tab:data} shows summary statistics for each dataset. 
We follow the preprocessing steps in~\cite{zhou-huang-2019-towards} by first replacing all numbers in both MWPs and equations to special tokens \texttt{num1}, \texttt{num2} etc. and then tokenizing both MWPs and equations into tokens and math symbols, respectively. In addition, we translate Math23K to English because this dataset is originally in Mandarin Chinese. Extension to languages other than English is left for future work.

Other popular MWP datasets such as Algebra~\cite{kushman-etal-2014-learning,upadhyay2015draw}, Dolphin18K~\cite{huang-etal-2016-well} and MathQA\footnote{MathQA is the most difficult  MWP dataset we have encountered, which containing GRE and GMAT level questions.}~\cite{amini-etal-2019-mathqa} contain MWPs with multiple equations and many variables, which are challenging to generate even for humans.
We leave the more challenging case of generating multi-variable, multi-equation MWPs to future work.

\paragraph{Setup and Baselines.}
We implement the LM and the mwp2eq models in our approach using pre-trained GPT-2~\cite{radford2019language}; one can also use other models since our approach is agnostic to the specific model architecture. 
We consider three baselines: {\bf seq2seq-rnn}, a sequence-to-sequence (seq2seq) model using LSTMs with attention that serves as the base architecture in~\cite{zhou-huang-2019-towards,2010.06196}; {\bf seq2seq-rnn-glove}, a modification to the previous baseline with GloVe~\cite{pennington-etal-2014-glove} instead of random embeddings at initialization; and {\bf seq2seq-tf}, a seq2seq model with transformers~\cite{NIPS2017_3f5ee243}. 
We also compare our approach to vanilla GPT-2, either randomly initialized or pre-trained; we denote these baselines as {\bf GPT} and {\bf GPT-pre}, respectively. 
For fair comparison, each baseline takes both equation and a set of keywords chosen by heuristics (see Section~\ref{sec:bsl}) as input to be consistent with the setup in our approach. 
For each dataset, we perform five-fold cross-validation and report the averaged evaluation results. See the Supplementary Material for more details on the experimental setup and baselines.

\paragraph{Metrics.}
For {language quality}, we use the following three evaluation metrics: {\bf BLEU-4}~\cite{Papineni:2002:BMA:1073083.1073135}, {\bf METEOR}~\cite{Lavie:2007:MAM:1626355.1626389}, and {\bf ROUGE-L}~\cite{lin:2004:ACLsummarization}, following recent literature on question generation~\cite{qg-net}. We implement these metrics using the package provided by~\cite{2015arXiv150400325C}.
For {mathematical consistency,}
We use the equation accuracy ({\bf ACC-eq}) metric that measures whether the generated MWP is mathematically consistent with the controlled input equation. The idea of this metric originates from other applications such as program translation and synthesis~\cite{10.5555/3327144.3327180,Chen2020Neural}.
In our case, because the equation associated with a generated MWP is not readily available, 
we resort to a mwp2eq model fine-tuned on each MWP dataset to predict the equation from an MWP. 
During the evaluation, we feed the generated MWP to the mwp2eq model as input and check whether the output of the mwp2eq model exactly matches the equation used as input to the MWP generator.

\begin{table*}[t!]
\centering
\caption{Generated MWP examples with fixed context and varying equations.}
\vspace{-5pt}
\label{tab:qual-context-eq}
\begin{adjustbox}{max width=0.9\textwidth}
\begin{tabular}{@{}p{9cm}p{9cm}@{}}
\toprule
\multicolumn{2}{c}{\bf Context: \texttt{candies}}                                                                                                                 \\ \midrule
\multicolumn{1}{c}{\bf Equation \#1: \texttt{x = num1 + num2}}                                                                                            & \multicolumn{1}{c}{\bf Equation \#2: \texttt{x = num1 - num2}}                                                                                                    \\\cmidrule(lr){1-1} \cmidrule(lr){2-2}
{\bf seq2seq-tf}: ethan has num1 presents . alissa has num2 more than ethan . how many presents does alissa have ?  ({\color{red}in training data})                      & {\bf seq2seq-tf}: mildred weighs num1 pounds . carol weighs num2 pounds . how much heavier is mildred than carol ?  ({\color{red}in training data})                               \\
{\bf GPT-pre}: There are num1 scissors in the drawer. Keith placed num2 scissors in the drawer. How many scissors are now there in total? ({\color{red} irrelevant to context})  & {\bf GPT-pre}: Joan has num1 blue balloons but lost num2 of them. How many blue balloons does Joan have now? ({\color{red} irrelevant to context})                                     \\
{\bf ours}: Mildred collects num1 candies. Mildred's father gives Mildred num2 more. How many candies does Mildred have?    ({\color{green}\cmark})              & {\bf ours}: There are num1 candies in the jar. num2 are eaten by a hippopotamus. How many candies are in the jar?     ({\color{green}\cmark})                            \\\midrule
\multicolumn{1}{c}{\bf Equation \#3: \texttt{x = num1 * num2}}                                                                                            & \multicolumn{1}{c}{\bf Equation \#4: \texttt{x = num1 / num2}}                                                                                                    \\\cmidrule(lr){1-1} \cmidrule(lr){2-2}
{\bf seq2seq-tf}: each banana costs \$ num1 . how much do num2 bananas cost ?  ({\color{red}in training data})                                                              & {\bf seq2seq-tf}: there are num1 bananas in diane ' s banana collection . if the bananas are organized into num2 groups , how big is each group ? ({\color{red}in training data})   \\
{\bf GPT-pre}: Joan has saved num1 quarters from washing cars. How many cents does Joan have? ({\color{red}inconsistent with equation})                                               & {\bf GPT-pre}: Joan has num1 blue marbles. Sandy has num2 times more blue marbles than Melanie. How many blue marbles does Joan have? ({\color{red}inconsistent with equation})              \\
{\bf ours}: Each child has num1 candies. If there are num2 children, how many candies are there in all? ({\color{green}\cmark})                                    & {\bf ours}: There are num1 candies in the candy collection. If the candies are organized into num2 groups, how big is each group? ({\color{green}\cmark})                  \\ \bottomrule
\end{tabular}
\end{adjustbox}
\vspace{-5pt}
\end{table*}

\begin{table}[]
\centering
\caption{Results of the ablation study, which validate the effectiveness of each component in our approach.}
\vspace{-8pt}
\label{tab:ablation}
\begin{adjustbox}{max width=\linewidth}
\begin{tabular}{@{}lcccccc@{}}
\toprule
                                & \multicolumn{2}{c}{\bf Arithmetic} & \multicolumn{2}{c}{\bf MAWPS} & \multicolumn{2}{c}{\bf Math23K} \\ \midrule
                                & BLEU-4         & ACC-eq        & BLEU-4      & ACC-eq      & BLEU-4       & ACC-eq       \\
                                \cmidrule(lr){2-3} \cmidrule(lr){4-5} \cmidrule(lr){6-7}
$\mathcal{L}_{\rm eq}$ (softmax)         & 0.110           & 0.417         & 0.308       & {\bf 0.555}       & 0.284        & 0.466              \\
{\bf $\mathcal{L}_{\rm eq}$ (Gumbel-softmax)}  & {\bf 0.303}          & {\bf 0.455}         & {\bf 0.522}       & {0.527}       & {\bf 0.306}        & {\bf 0.495}        \\\hline
keyword, TF-IDF                 & 0.313                & {\bf 0.424}             & 0.518            & 0.536            & 0.310             & 0.498             \\
keyword, noun+pronoun           &  0.316              & 0.413              &  0.504           & 0.512            & {\bf 0.325}             & 0.498             \\
{\bf context selection}         & {\bf 0.320}          & 0.412         & {\bf 0.533}       & {\bf 0.542}       & {0.324}        & {\bf 0.501}        \\\hline
full model w/o $\mathcal{L}_{\rm c}$ & 0.303          & {\bf 0.455}         & 0.522       & 0.527       & 0.306        & 0.495        \\
full model w/o $\mathcal{L}_{\rm eq}$    & 0.320          & 0.412         & 0.491       & 0.500       & 0.324        & 0.501        \\
full model w/o both             & 0.316          & 0.403         & 0.504       & 0.512       & 0.325        & 0.498        \\
{\bf full model}                & {\bf 0.338}    & 0.453         & {\bf 0.596} & {\bf 0.557} & {\bf 0.332}  & {\bf 0.513}  \\
\bottomrule
\end{tabular}
\end{adjustbox}
\vspace{-10pt}
\end{table}

\subsection{Quantitative Results}
\label{sec:quant}
Table~\ref{tab:quant-results} shows the quantitative results of our experiments.
The number in parenthesis below ACC-eq is the equation accuracy when we feed the mwp2eq model the ground-truth MWPs in the respective datasets.
We see that our approach outperforms the best baseline on most occasions, especially on language quality metrics. 
However, there are a few exceptions, especially for the ACC-eq metric on the Math23K dataset. Specifically, we note that the seq2seq-tf baseline seems to yield an ACC-eq value even higher than the oracle accuracy at the first attempt. 
Upon closer investigation, we find that the baseline seq2seq models, especially the seq2seq-tf baseline, simply memorize the training data.  
Table~\ref{tab:unique} illustrates this finding and shows the percentage of generated MWPs that are {\em not} present in the training data. 
We see that the seq2seq-tf baseline tends to directly copy MWPs from the training data as its ``generated'' MWPs, especially on the 2 smaller datasets. 
In contrast, our approach generates novel MWPs most of the time. 
We thus report ACC-eq only on the novel MWPs generated by the seq2seq-tf baseline on the Math23K dataset.
Our approach outperforms seq2seq-tf on this modified ACC-eq metric. 

\begin{table*}[t!]
\centering
\caption{Generated MWP examples with novel context not present in the training data.}
\vspace{-5pt}
\label{tab:qual-eq-context}
\begin{adjustbox}{max width=0.9\textwidth}
\begin{tabular}{@{}p{9cm}p{9cm}@{}}
\toprule
\multicolumn{2}{c}{\bf Equation: \texttt{x = num1 + num2 + num3}}                                                                                                                 \\ \midrule
\multicolumn{1}{c}{\bf Context \#1: \texttt{violin piano acoustic guitar}}                                                                                            & \multicolumn{1}{c}{\bf Context \#2: \texttt{beets eggplant}}                                                                                                    \\\cmidrule(lr){1-1} \cmidrule(lr){2-2}
{\bf seq2seq-tf}: sara grew num1 onions , sally grew num2 onions , and fred grew num3 onions . how many onions did they grow in all ?
  ({\color{red}in training data})                                                              
  & {\bf seq2seq-tf}: sara grew num1 onions , sally grew num2 onions , and fred grew num3 onions . how many onions did they grow in all ? ({\color{red}in training data}) 
  \\
{\bf GPT-pre}: There are num1 dogwood trees currently in the park. Park workers will plant num2 dogwood trees today and num3 dogwood trees tomorrow. How many dogwood trees will the park have when the workers are finished? ({\color{red}irrelevant to context})                                               & {\bf GPT-pre}: There are num1 orchid bushes currently in the park. Park workers will plant num2 orchid bushes today and num3 orchid bushes tomorrow. How many orchid bushes will the park have when the workers are finished? ({\color{red}irrelevant to context})            \\
{\bf ours}: Mike joined his school's band. He bought a clarinet for \$ num1, a music stand for \$ num2, and a song book for \$ num3. How much did Mike spend at the music store? ({\color{green}\cmark})                                    & {\bf ours}: Sara grew num1 beets, Sally grew num2 beets, and Fred grew num3 beets. How many beets did they grow in total? ({\color{green}\cmark})                 \\ \bottomrule
\end{tabular}
\end{adjustbox}
\vspace{-5pt}
\end{table*}

\paragraph{Ablation Study.} 
To validate that each component in our approach contributes to its success, we conduct an ablation study and compare our approach with several variants and several baselines after removing some of these components. 
For the use of Gumbel-softmax in the equation consistency constraint computation, we compare to softmax \cite{dlbook}, which removes sampling from the Gumbel variable. 
For the context keyword selection model, we compare to several context keyword selection heuristics including TF-IDF \cite{tfidf,zhou-huang-2019-towards} and nouns+pronouns; see the Supplementary Material for more details on these baselines.  
Table~\ref{tab:ablation} shows the ablation study results, reporting on BLEU-4 and ACC-eq as the representative metric for language quality and mathematical consistency, respectively. These comparisons validate the necessity of each component in our approach: Gumbel-softmax outperforms softmax and our context keyword selection method outperforms other heuristic methods. 
We also see that our approach outperforms variants with either component removed and that the equation consistency constraint and the context keyword selection method tend to improve the mathematical consistency and language quality of the generated MWPs, respectively.

\subsection{Qualitative Results}
\label{sec:qual}
Since seq2seq baselines outperform our approach on a few occasions under the automated metrics, we now conduct a few case studies to investigate each approach. We investigate i) how controllable is each approach by giving it different input equations and contexts and ii) how generalizable each approach is by giving it unseen contexts in the dataset. 
Specifically, we conduct two qualitative experiments: First, we hold an input context fixed and change the input equation; Second, we hold the input equation fixed and change the context. 
We compare the MWPs generated by our approach to those generated by the seq2seq-tf and GPT-pre baselines trained on the MAWPS dataset, where these baselines perform well under automated metrics (see Table~\ref{tab:quant-results}).
The Supplementary Material contains additional qualitative examples.

\paragraph{Fixed Context, Changing Equation.}
Table~\ref{tab:qual-context-eq} shows the MWPs generated by each approach using the same input context and different input equations. 
We see that every approach can generate MWPs with high language quality and are mathematically valid most of the time. 
However, upon closer inspection, we find that MWPs generated by the seq2seq-tf baseline are often exact copies of those it has seen in the training data. 
In other words, the model does nothing more than memorizing the training data and retrieving the most relevant one given the input equation and context; see Table~\ref{tab:unique} for a numeric comparison and the discussion in Section~\ref{sec:quant}.
This observation is not surprising because training only on small MWP datasets leads to overfitting. 
It also explains why the seq2seq baselines perform well on the automated metrics since these MWP datasets contain problems that lack language diversity, which results in many overlapping words and phrases that often appear in both the training and validation sets. The GPT-pre baseline, on the other hand, is sometimes capable of generating novel MWPs, but they are either irrelevant to the input context or are inconsistent with the input equation. 
Only our approach consistently generates MWPs that are both novel and mathematically consistent with the input equation.

\paragraph{Fixed Equation, Changing Context.}
Table~\ref{tab:qual-eq-context} shows the MWPs generated by each approach using the same input equation and different input contexts. 
The keywords in these contexts are not part of the vocabulary of the training set and are thus unseen by the model during training/fine-tuning. 
Similar to the results of the previous experiment, here we also see that the seq2seq baseline simply retrieves an MWP from the training dataset as its ``generated'' MWP. 
This observation is unsurprising for the seq2seq baseline because it simply converts an out-of-vocabulary word in the input context into a special \texttt{unknown} token, which is uninformative. 
Interestingly, the GPT-pre baseline also generates MWPs that have minimal difference from MWPs in the training set or seems to ignore the input context. We again attribute this phenomenon to the small dataset size, on which the model also overfits if no additional constraints are introduced.
Once again, in this setting, only our approach consistently generates novel and high-quality MWPs that are relevant to the input context.

\begin{table}[]
\centering
\caption{Examples of the keywords that are selected from a (possibly long) input context.}
\vspace{-5pt}
\label{tab:kw}
\begin{adjustbox}{max width=0.9\linewidth}
\begin{tabular}{@{}p{8cm}@{}}
\toprule
{\bf MWP:} Emily collects num1 cards . Emily ' s father gives Emily num2 more . Bruce has apples . How many cards does Emily have ?                              \\ 
{\bf Context keywords:} Emily cards collects father                                                                                                                           \\\midrule
{\bf MWP:} The school cafeteria had num1 apples . If they used num3 to make lunch for the students and then bought num2 more , how many apples would they have ? \\
{\bf Context keywords:} apples cafeteria                                                          
                                                                                                                                                      \\ \bottomrule
\end{tabular}
\end{adjustbox}
\vspace{-5pt}
\end{table}

\paragraph{Selected Context Keywords.}
To investigate our context keyword selection model, we show in Table~\ref{tab:kw} a few examples of the input context (which is the original MWP in our training setting) and the selected context keywords, i.e., those with $c^{(i)} > 0.5$ (recall Eq.~\ref{eq:c}). 
We see that our context keyword selection model can identify components that are key to the relevant underlying mathematical components in the MWP; for example, it identifies only ``Emily collects cards father'' as the key to this MWP and ignores the part with ``Bruce apples'', which is unrelated to the math equation. Such a context keyword selection method is useful in practice to summarize (possibly long) input contexts provided by human instructors/content designers.

\section{Related Work}
\label{sec:rw}

\paragraph{MWP Generation and Answering.}
Earlier works on MWP generation do so in a highly structured way, explicitly relying on domain knowledge and even pre-defined equation and text templates~\cite{Nandhini2011,DBLP:conf/aaaifs/Williams11,10.5555/2832249.2832302,deane2003automatic}. 
More recently, neural network-based approaches have shown significant advantages in generating high-quality questions compared to template-based approaches. 
Recent approaches on MWP generation also take this approach, usually using recurrent neural networks in a seq2seq pipeline~\cite{zhou-huang-2019-towards,2010.06196}. 
Instead of focusing on building new datasets or specific model architectures, we tackle the MWP generation problem from a controllable generation perspective, where we focus on the generated MWPs' language quality and mathematical consistency. 
This focus leads to our proposed approach that specifically aims at tackling these two challenges; our framework is model-agnostic and can be combined with almost any existing MWP generation approach.

Our approach also involves a model (mwp2eq) that parses an MWP into its underlying equation, which has been a very active research area with a plethora of related work, e.g.,~\cite{huang-etal-2018-neural,chiang-chen-2019-semantically,Xie2019,zou-lu-2019-text2math,li-etal-2019-modeling,li-etal-2020-graph-tree,qin-etal-2020-semantically,shi-etal-2015-automatically,DBLP:conf/aaai/WangZGSGS18,wang-etal-2018-translating,roy-roth-2015-solving,Wang2019,amini-etal-2019-mathqa,wu-etal-2020-knowledge}. In this work, we simply use a pre-trained LM as the mwp2eq model; investigation of leveraging the above recent advances to improve mathematical consistency of the generated MWPs is left for future work.

\paragraph{Controllable Text Generation.}
Our work is also related to a growing body of literature on controllable text generation~\cite{prabhumoye-etal-2020-exploring,wang-etal-2019-topic-guided,10.5555/3305381.3305545,keskarCTRL2019,shen-etal-2019-select}. 
In particular, our equation consistency constraint takes inspiration from the above works that impose similar constraints to improve control over the generation process.
A major difference between our work and most of these prior works is that,
in most of these approaches, the control elements, such as emotion, sentiment, and speaker identity, are usually represented as scalar numerical values. In contrast, our control elements (equation and context) consist of a sequence of math symbols or tokens rather than numeric values, which requires additional technical solutions to propagate gradient.
The Gumbel-softmax trick~\cite{gumbel1,gumbel2} that we employ has found success in text generation using generative adversarial networks (GANs)~\cite{1611.04051,10.5555/3327345.3327377,nie2018relgan,NEURIPS2020_3eb46aa5,Jiao2021}, a setting similar to ours where discrete sampling becomes an issue.

\section{Conclusions and Future Work}
In this paper, we 
developed a controllable MWP generation approach that (i) leverages pre-trained language models to improve language quality of the generated MWP, (ii) imposes an equation consistency constraint to improve mathematical consistency of the generated MWP, and (iii) includes a context selector that sets the context (in the form of a set of keywords) to use in the generation process. Experimental results on several real-world MWP datasets show that, while there is plenty of room for improvement, our approach outperforms existing approaches at generating mathematically consistent MWPs with high language quality.

Automatically generating MWPs remains a challenging problem and our work opens up many avenues for future work. 
First, our study is limited to the case of simple MWPs, each with a single equation and variable.
While results are encouraging, our approach does not generalize well to the more challenging case when the input consists of multiple, complex equations. 
In these cases, we need more informative representations of the input equations~\cite{forte}. 
Second, there is no clear metric that can be used to evaluate the generated MWPs, especially their mathematical validity. 
It is not uncommon when a generated MWP with high scores under our metrics is either unanswerable or inconsistent with the input equation. 
Therefore, future work should also focus on developing metrics for better evaluation of generated MWPs' mathematical validity. %
Last but not least, while we focus on 2 control elements (equation, context), an interesting future direction is to add more control elements to the generation process such as question difficulty and linguistic complexity.

\section*{Acknowledgements}
This work is supported by NSF grants 1842378, 1937134, and 1917713, ONR grant N0014-20-1-2534, AFOSR grant FA9550-18-1-0478, and a Vannevar Bush Faculty Fellowship, ONR grant N00014-18-1-2047.

\bibliography{emnlp2021}
\bibliographystyle{acl_natbib}

\clearpage
\newpage
\setcounter{section}{0}
\setcounter{figure}{0}
\setcounter{table}{0}
\renewcommand{\thesection}{\Alph{section}}
\renewcommand{\thetable}{\Alph{table}}
\renewcommand{\thefigure}{\Alph{figure}}

\begin{center}
{\LARGE \bf
Supplementary Material \\[0.15cm] for}\\[0.15cm]
{\large\bf{Math Word Problem Generation with Mathematical Consistency and Problem Context Constraints}}

\end{center}

\section{Gumbel-Softmax in Section~\ref{sec:eq}}
We describe in detail the procedure to approximate sampling $M'$ from $p_\Phi$, i.e., sampling discrete tokens $m_t' \sim p_\Phi (m_t | E, \rvc, m_1,\ldots, m_{t-1})$, using the Gumbel-softmax relaxation. 
In the first step, we reparametrize sampling from a categorical distribution $p_\Theta$ using the Gumbel-max trick~\cite{gumbel2} as follows:
\begin{align}
    u^{(i)} &\sim {\rm uniform}(0,1)\,,\nonumber \\
    g_t^{(i)} &= -{\rm log} (-{\rm log} (u^{(i)}))\,,\nonumber \\
    m_t &= {\rm one\_hot} \bigg(\underset{i\in |V|}{\rm argmax} \big(f_{\Theta,t}^{(i)} + g_t^{(i)}\big) \bigg)\,,\nonumber 
\end{align}
where $|V|$ is the size of the vocabulary, $f_{\Theta,t}^{(i)}$ is the pre-softmax activation of $p_\Theta$ at the $t$-th generation step for the $i$-th word, and $g_t^{(i)}$ are i.i.d.\ samples from the standard Gumbel distribution. 

In the second step, we approximate the discrete argmax operator with the continuous, differentiable softmax operator, which enables us to obtain the final approximation
\begin{align}
    m_t' = {\rm softmax}((f_{\Theta,t} + g_t)/\tau)\,,\nonumber
\end{align}
where $\tau$ is a temperature hyperparameter, resulting in the Gumbel-softmax distribution. When $\tau$ approaches $0$, this approximation approaches the categorical distribution parametrized by ${\rm one\_hot} \big({\rm argmax}_{i\in|V|} (f_{\Theta,t})\big)$.

\begin{table*}[]
\caption{Model configurations.}
\vspace{-5pt}
\label{tab:exp-setting-detail}
\begin{adjustbox}{max width=\textwidth}
\begin{tabular}{@{}lcccccccc@{}}
\toprule
\textbf{architecture} & \textbf{\#layers} & \textbf{input size} & \textbf{layer size} & \textbf{\#params} & \textbf{optimizer} & \textbf{learning rate} & \textbf{batch size} & \textbf{training epochs/steps} \\ \midrule
seq2seq-rnn           & 2                 & 300                 & 512                 & 11M               & adagrad            & 0.15                   & 64                  & \{5000, 15000$^*$\}                      \\
seq2seq-attn-rnn      & 1                 & 300                 & 512                 & 11M               & adagrad            & 0.15                   & 64                  & \{5000, 15000$^*$\}                      \\
seq2seq-transformer   & 6                 & 512                 & 512                 & 52M               & Adam               & 2                      & 4096                & \{5000, 15000$^*$\}                      \\
GPT                   & 36                & 1280                & 1280                & 774M              & Adam               & 5e-5                   & 8                   & 20                       \\
ours                  & 36                & 1280                & 1280                & 774M              & Adam               & 5e-5     & \{8$^*$, 16\}          & \{5, 15$^\dagger$\}              \\ \bottomrule
\end{tabular}
\end{adjustbox}
\end{table*}

\begin{table*}[t!]
\caption{Additional examples of MWPs generated by our approach.}
\vspace{-5pt}
\label{tab:qual-more}
\centering
\begin{adjustbox}{max width=0.9\textwidth}
\begin{tabular}{p{2cm}p{0.8\linewidth}}
\toprule
{\bf Equation} & \texttt{x = num1 / num2}
\\
{\bf Context} & \texttt{Sue bag bags cookies fill mother}
\\
{\bf Gen. MWP} & Sue ’ s mom baked num1 cookies. If she wants to distribute the cookies among num2 children, how many cookies will each child get?
\\\midrule
{\bf Equation} & \texttt{x = num1 + num2 + num3}
\\
{\bf Context} & \texttt{Charlie Cortland bag bags fruit pick picked visit}
\\
{\bf Gen. MWP} & Cortland picked num1 pears, and Mike picked num2 pears, and Alyssa picked num3 pears from the pear tree. How many pears were picked in total?
\\\midrule
{\bf Equation} & \texttt{x = num1 - num2}
\\
{\bf Context} & \texttt{cousin game playing points scored video
}
\\
{\bf Gen. MWP} & Zach scored num1 points in the football game. Ben scored num2 points. How many more points did Zach score than Ben?
\\\midrule
{\bf Equation} & \texttt{x = num1 + num2}
\\
{\bf Context} & \texttt{teacher worksheet}
\\
{\bf Gen. MWP} & The secretary prints a copy of the worksheet, num1 copies in the morning, and num2 copies in the afternoon. How many copies were printed throughout the day?
\\ \midrule
{\bf Equation} & \texttt{x = ( num1 + num2 )}
\\ 
{\bf Context} & \texttt{mike baseball football marble total toy}
\\ 
{\bf Gen. MWP} & The total cost of a toy factory to produce a football is num1 yuan, which is num2 yuan less than the total cost. How much is the total cost?
\\\midrule
{\bf Equation} & \texttt{x = ( num1 * num2 )}\\
{\bf Context} & \texttt{anne hour mile} \\
{\bf Gen. MWP} & It takes num2 hours for a car to travel num1 kilometers per hour from A to B. How many kilometers are the distance between A and B? \\

 \bottomrule

\end{tabular}
\end{adjustbox}
\end{table*}

\section{Quality of the Math23K Dataset}
\begin{figure}
    \centering
    \includegraphics[width=0.8\linewidth]{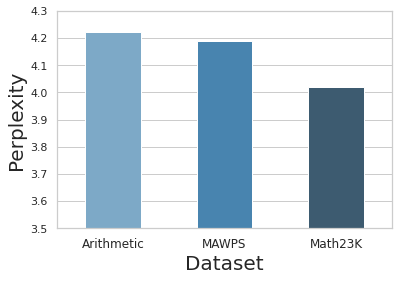}
    \vspace{-8pt}
    \caption{Averaged perplexity of each dataset under a small GPT-2. The translated Math23K dataset has similar perplexity compared to the other two datasets, suggesting similar language quality of the three datasets.}
    \label{fig:dataset_ppl}
    \vspace{-5pt}
\end{figure}

Some reviewers brought up a concern on the quality of the Math23K dataset because it is originally in Chinese and we use the English-translated version (via Google Translate API) of this dataset. Despite using an automated translation service, we find that most data points in the translated Math23K dataset is good enough to use for training and evaluation. Figure~\ref{fig:dataset_ppl} reports the perplexity score under a small GPT-2 model for each dataset, averaged over all data points. We see that the translated Math23K dataset has comparable perplexity compared to that of the other two datasets. This observation suggests that the translated Math23K dataset has similar language quality compared to the other two datasets that are originally in English. 

\section{Experiment Details}

\subsection{Training Details}
We train the three components in our method jointly. Specifically, we first jointly train the context keyword selection model and the MWP generator. After that, we freeze the context selection model and continue to jointly train the MWP generator and the mwp2eq model. Notably, the input token embeddings to the context selector are the same as the pre-trained GPT-2 token embeddings. These embeddings are kept fixed throughout training for training stability.

Table~\ref{tab:exp-setting-detail} provides the configurations for all models under consideration. The numbers marked with an asteroid (*) are the setting for the Math23K dataset. The number marked with a dagger ($\dagger$) is the configuration for our approach. For all baselines, we use the noun+pronoun words (see Section~\ref{sec:bsl}) extracted from the MWPs as the input context. Each model are trained on a single NVIDIA RTX 8000 GPU. For the GPT-based models, including our approach, the training time ranging from 1.3 minute per epoch (for the Arithmetic and MAWPS datasets) to 20 min per epoch (for the Math23K dataset). The much slower runtime for the Math23K dataset is due to both its smaller batch size in order to fit into memory and its large size, which is almost 20 times as big as the Arithmetic dataset. For the seq2seq baselines, the training speed is about 7 seconds per 10 steps (equivalent of 640 data points). The seq2seq baselines are implemented using the OpenNMT framework.\footnote{\url{https://opennmt.net/}} The pre-trained GPT-2 model is loaded from the HuggingFace repository.\footnote{\url{https://huggingface.co/}}

When training all models, we also randomly drop words in the context with a probability of 0.3 for each word and permute the order of these words. We do so to improve the models' robustness with respect to the number and the order of the keywords in the context.
In addition, for the seq2seq baselines, we additionally lowercase all string to reduce the total number of vocabulary.

\subsection{Baselines for the Ablation Study}
\label{sec:bsl}
Here, we explain the different method to select keywords. We first tokenize\footnote{\url{https://spacy.io/}} the MWPs. For the noun+pronoun method, we extract words that have ``noun'' or ``pronoun'' as their part-of-speech tags. Stopwords and punctuation are not included. For the TF-IDF method, we compute the TF-IDF weights for all tokens, again excluding stopwords and punctuation, and then choose 5 words with the highest weights for each MWP as its context.

\begin{table*}[h!]
\centering
\caption{Examples of failed cases.}
\label{tab:error-analysis}
\vspace{-5pt}
\begin{adjustbox}{max width=0.9\textwidth}
\begin{tabular}{p{2cm}p{0.88\linewidth}}
\toprule
{\bf Equation}      & \texttt{x = ( num1 / num2)}                                                                                                                                                                                                                                                                  \\
{\bf Context}       & \texttt{Tom cars dollars money week weekend}                                                                                                                                                                                                                                                   \\
{\bf Gen. MWP} & On Sunday, num1 yuan can buy num2 cars. So how much money is needed to buy a car? \\
{\bf\leavevmode\color{red}Error} & {\leavevmode\color{red}Incomplete information: implicitly assumes each car costs the same.} Can be remedied by adding ``on average''.) \\ \midrule
{\bf Equation} & \texttt{x = ( num1 * num2 )} \\
{\bf Context} & \texttt{Tom box boxes brother candy chocolate pieces} \\
{\bf Gen. MWP} & There are num2 boxes of chocolates in the candy store, and the price is num1 yuan per piece. How much does it cost to buy a piece of chocolate? \\ 
{\bf\leavevmode\color{red}Error} & {\leavevmode\color{red}Wrong question asked.} Can be remedied by changing the ``a piece of chocolate'' to ``those chocolates''. \\ \midrule
{\bf Equation}      & \texttt{x = ( num1 * num2 )}                                                                                                                                                                                                                                        \\
{\bf Context}       & \texttt{David box dog dogs dollars toy}                                                                                                                                                                                                                                                           \\
{\bf Gen. MWP} & A toy dog is num1 yuan, and the price of a puppy is num2 times that of a puppy. How much is a puppy? \\
{\bf\leavevmode\color{red} Error} & {\leavevmode\color{red}Incoherent question statement.} Can be remedied by changing the second ``puppy'' to ``dog''.                                                                                                                                                       \\ \bottomrule
\end{tabular}
\end{adjustbox}
\end{table*}

\subsection{Mathematical Consistency Metric}
In principle, A more accurate mwp2eq model leads to more accurate mathematical consistency evaluation and many other state-of-the-art mwp2eq methods, including those targeting automatic MWP answering that we reviewed in Section~\ref{sec:rw}, can be employed. We have observed that fine-tuning pre-trained GPT-2 achieves competitive performance comparing to a number of existing approaches and thus use it for this present work.
Using more advanced methods to improve the mathematical consistency evaluation is left for future work.

\section{Additional Results}
\begin{figure}
    \centering
    \includegraphics[width=0.8\linewidth]{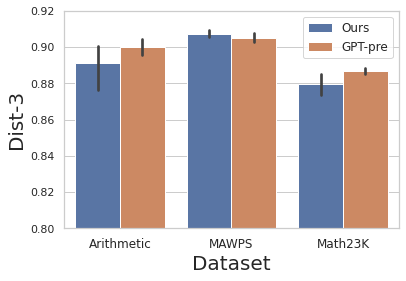}
    \caption{Diversity of generation comparing our approach with a fine-tuned pre-trained GPT-2. Our approach achieves similar generation diversity according to the Dist-3 metric.}
    \label{fig:diversity}
\end{figure}

\subsection{Generation Diversity}
Per the reviewers request, we compare the generation diversity using the Dist-3 metric~\cite{li-etal-2016-diversity} in Figure~\ref{fig:diversity}, where higher numbers indicating more diversity. We can observe that our approach achieves similar generation diversity across all datasets compared to GPT-2, with differences smaller than 0.1, suggesting our regularizations do not compromise the generation diversity.

\subsection{Additional Qualitative Examples}
Table~\ref{tab:qual-more} presents additional examples of MWPs generated by our approach. The contexts and equations in the first three rows and the last three rows are taken from the Arithmetic and the Math23K datasets, respectively. 
These examples are consistent with the qualitative results in Section~\ref{sec:qual}.

\section{Limitations}
\label{sec:limitation}
Despite promising results, our approach can still generates problematic MWPs. 
Because some of our baselines simply copy a sample from the training data as the ``generated'' sample during evaluation, which would make a unfair comparison, here we instead conduct a small case study for our approach on the most challenging generation scenario where we randomly sample 25 contexts and combine it with each of the four equations that involve two variables.\footnote{In general, evaluating generated MWPs is a challenging task and we defer the investigation of human evaluation criteria and more comprehensive human evaluations to a future work.} This procedure produces 100 generated examples. We then qualitatively evaluate their generation quality. In total, we find that 17 out of the 100 generated samples are completely satisfactory and another 17 can become satisfactory with minor changes. Some common errors in our generated samples include: 1) incomplete information; 2) wrong question asked; and 3) incoherent question statement. We illustrate these types of errors in Table~\ref{tab:error-analysis}. 
The errors suggest that, for better generation quality, we should further improve the model's understanding of the semantics of the math operations and the relationship between various mathematical entities in the equation and the important words in the MWPs.

\end{document}